# Model-based Utility Functions

**Bill Hibbard**
TEST@SSEC.WISC.EDU
*Space Science and Engineering Center*
*University of Wisconsin - Madison*
*1225 W. Dayton Street*
*Madison, WI 53706, USA*

## Abstract

Orseau and Ring, as well as Dewey, have recently described problems, including self-delusion, with the behavior of agents using various definitions of utility functions. An agent's utility function is defined in terms of the agent's history of interactions with its environment. This paper argues, via two examples, that the behavior problems can be avoided by formulating the utility function in two steps: 1) inferring a model of the environment from interactions, and 2) computing utility as a function of the environment model. Basing a utility function on a model that the agent must learn implies that the utility function must initially be expressed in terms of specifications to be matched to structures in the learned model. These specifications constitute prior assumptions about the environment so this approach will not work with arbitrary environments. But the approach should work for agents designed by humans to act in the physical world. The paper also addresses the issue of self-modifying agents and shows that if provided with the possibility to modify their utility functions agents will not choose to do so, under some usual assumptions.

## 1. Introduction

Hutter significantly advanced the mathematical theory of rational agents by his definition of the agent AIXI (AI for Artificial Intelligence and XI for the Greek letter $\xi$, whose role is explained in Section 2) and proof that it is optimal but uncomputable (2005). Hutter's work uses a mathematical framework that allows properties of agents to be proved. In this framework an agent's motives are defined by a utility function to be maximized. Omohundro discussed the drives of artificial intelligence (AI) agents in a closely reasoned but non-mathematical paper (2008). He argued that AI agents will try to preserve their utility functions and prevent them from corruption by other agents.

More recently, in mathematical papers, Orseau and Ring (2011a; 2011b) and Dewey (2011) have argued that under broad assumptions about utility functions, agents will behave in ways not anticipated or wanted by their designers. Dewey argues that artificial agents whose utility functions are designed by humans will modify their environments so that they can maximize their utility functions without accomplishing the intentions of human designers. Ring and Orseau argue that agents will intentionally alter their own observations of their environment, that is self-delude, to maximize their utility function regardless of the actual state of the environment.

This paper describes an approach to addressing these problems. In the problems described by Dewey and by Ring and Orseau for reinforcement learning agents, utility functions are defined in terms of the agent's observations of the environment. In my solution approach utility functions are defined in terms of a model of the environment that the agent must learn via its interactions with that environment. A critical difference is that such a model-based utility function must be defined in terms of both an agent's observations of the environment and its actions on the environment, since the environment model is learned from both observations and actions. Agents with model-based utility functions are a special case of model-based agents, which include any agents with environment models.

As humans design increasingly complex AI agents, those agents will need to learn their world models rather than having pre-programmed world models. And rather than their actions being pre-programmed, the agents will need utility functions to motivate their actions. The contribution of this paper is to demonstrate how utility functions can be defined so that agents do not self-delude. A limitation of this paper is that the demonstration is by examples rather than a proof that some broad class of agents do not self-delude.

The solution approach presented in this paper adds an unavoidable complexity to the definition of utility functions: they are defined in terms of an environment model that must be learned and hence is unknown at the time the agent is defined. This requires that utility function definitions be based on some assumptions about the environment. For example an agent designer may know that the environment contains certain kinds of objects and define the utility function in terms of the states of those objects. The utility function may initially be defined in terms of specifications that will match those objects and their states in the learned environment model. The specifications in an agent's utility function definition may fail to match any structures in the agent's learned environment model, in which case the agent fails (or continues to search its environment for structures that do match the specifications). This approach to solving the problem of self-delusion will not work for arbitrary environments, but it should work for agents designed to act in the physical world. It is common to define model-based utility functions for agents with preprogrammed environment models, but the proposal of model-based utility functions for agents that learn their environment models is novel. One previous reference is my prescription for the utility functions of possible future super-intelligent machines, which specifies that they initially learn and constantly relearn to recognize humans and their emotions (Hibbard, 2008).

The next section of this paper describes the mathematical framework for reasoning about agents and environments. The third section describes self-delusion and other problems with agent behavior. The fourth section presents the model-based approach to defining utility functions, illustrated by two simple examples. The fifth section discusses the possibility that an agent may modify itself so that it self-deludes, and proves that this will not happen under certain assumptions. The concluding section discusses the paper's results and their significance.

## 2. An Agent-Environment Framework

We assume that an agent interacts with an environment. At each of a discrete series of time steps $t \in \mathbf{N} = \{0, 1, 2, ...\}$ the agent sends an action $a_t \in A$ to the environment and receives an observation $o_t \in O$ from the environment, where $A$ and $O$ are finite sets. We assume that the environment is computable and we model it by programs $q \in Q$, where $Q$ is the set of programs for a deterministic prefix universal Turing machine (PUTM) $U$ (Li and Vitanyi, 1997; Hutter, 2005). The environment may be non-deterministic in which case it is modeled by a distribution of deterministic programs. Let $h = (a_1, o_1, ..., a_t, o_t) \in H$ be an interaction history where $H$ is the set of all finite histories, and define $|h| = t$ as the length of the history $h$. Given a PUTM program $q \in Q$ we write $o(h) = U(q, a(h))$, where $o(h) = (o_1, ..., o_t)$ and $a(h) = (a_1, ..., a_t)$, to mean that $q$ produces the observations $o_i$ in response to the actions $a_i$ for $1 \leq i \leq t$. Given a program $q$ the probability $\rho(q) : Q \rightarrow [0, 1]$ is the agent's prior belief that $q$ is a true model of the environment. The prior probability of history $h$ is given by:

$$\rho(h) = \sum_{q:o(h)=U(q,\, a(h))} \rho(q) \tag{1}$$

An agent is motivated according to a *utility function* $u : H \rightarrow [0, 1]$ which assigns utilities between 0 and 1 to histories. Future utilities are discounted according to a *temporal discount function* $w : \mathbf{N} \rightarrow [0, 1]$ such that $\sum_{t\in\mathbf{N}} w(t) < \infty$. This is often chosen as geometric discounting $w(t) = \gamma^t$ where $0 < \gamma < 1$ (Sutton and Barto, 1998). The value $v_t(h)$ of a possible future history $h$ at time $t$ is defined recursively by:

$$v_t(h) = w(|h| - t)\, u(h) + \max_{a\in A} v_t(ha) \tag{2}$$
$$v_t(ha) = \sum_{o\in O} \rho(o \mid ha)\, v_t(hao)$$

Then the agent, denoted $\Pi(\rho, u, w, A, O)$, is defined to take, after history $h$, the action:

$$a_{|h|+1} = \text{argmax}_{a\in A}\, v_{|h|+1}(ha)$$

Denote the length of the PUTM program $q$ as $|q|$ (this length would be a count of bits for a binary program). Hutter (2005) defined the agent AIXI as $\Pi(\xi, u, w, A, O)$ where $\xi(h)$, AIXI's prior probability for history $h$, is derived from $\xi(q) = 2^{-|q|}$ by substituting $\xi$ for $\rho$ in (1), and where the utility function is defined from a reward as described in the next section. This is related to the Kolmogorov complexity $K(h)$ of a history $h$, which is defined as the length of the shortest program $q$ such that $o(h) = U(q, a(h))$. Kraft's Inequality implies that $\xi(h) \leq 1$ (Li and Vitanyi, 1997). Hutter showed that AIXI maximizes the expected value of future history, although AIXI is not computable.

## 3. Self-delusion

Orseau and Ring (2011a; 2011b) defined four types of $\Pi(\rho, u, w, A, O)$ agents using different specialized ways of computing utility $u(h)$ and temporal discount $w(t)$. These were then analyzed with respect to possible behaviors including self-delusion and self-modification.

For a *reinforcement-learning agent*, the observation includes a reward $r_t$ (i.e., $o_t = (\hat{o}_t, r_t)$) and $u(h) = r_{|h|}$. The temporal discount function is $w(t) = 1$ if $t \leq m$, where $m$ is a constant future horizon, and 0 otherwise.

For a *goal-seeking agent*, $u(h) = g(o_1, ..., o_{|h|}) = 1$ if the goal is reached at time $|h|$ and is 0 otherwise, where the goal can be reached at most once in any history. The temporal discount function is $w(t) = 2^{-t}$.

For a *prediction-seeking agent*, $u(h) = 1$ if $o_{|h|+1} = \text{argmax}_{o\in O}\, \rho(o \mid ha)$ and 0 otherwise, where $a$ is the agent's action after $h$ (that is, $u(h) = 1$ when the agent correctly predicts its next observation). The temporal discount function is the same as for the reinforcement-learning agent.

For a *knowledge-seeking agent*, $u(h) = -\rho(h)$ and $w(t) = 1$ if $t = m$, where $m$ is a constant, and 0 otherwise.

Ring and Orseau (2011b) defined a *delusion box* that an agent may choose to use to modify the observations it receives from the environment, in order to get the "illusion" of maximal utility (maximal reward or quickest path to reaching the goal). The delusion box is expressed as a function $d : O \rightarrow O$ that modifies the agent's observations. The code to implement $d$ is set as part of the agent's action $a = (a_e, d)$ sent to the environment, as illustrated in Figure 1. The agent receives observation $o = d(o_e)$ transformed by the delusion box from the observation $o_e$ sent by the environment. In their Statements 1, 2 and 3, Ring and Orseau argue (their arguments are not complete proofs though they are written formally) that reinforcement-learning, goal-seeking and prediction-seeking agents will all choose to use the delusion box. Their arguments use $P(\text{DB})$, the agent's estimate of the probability that the delusion box exists. In the reinforcement-learning case they argue that the agent will get constant reward $r_{\text{DB}} = 1$ using the delusion box and will get expected average reward $r_{\neg\text{DB}} < 1$ not using the delusion box. The agent's expected value choosing to use the delusion box is $v_{\text{DB}}(h) \geq r_{\text{DB}}\ P(\text{DB}) = P(\text{DB})$ and its expected value not choosing to use the delusion box is $v_{\neg\text{DB}}(h) \leq r_{\neg\text{DB}}\ P(\text{DB}) + (1 - P(\text{DB}))$. As the agent explores its environment, it can increase $P(\text{DB})$ arbitrarily close to 1 so that $v_{\text{DB}}(h) > v_{\neg\text{DB}}(h)$ and the agent will choose to use the delusion box. They make a related argument in the goal-seeking case.

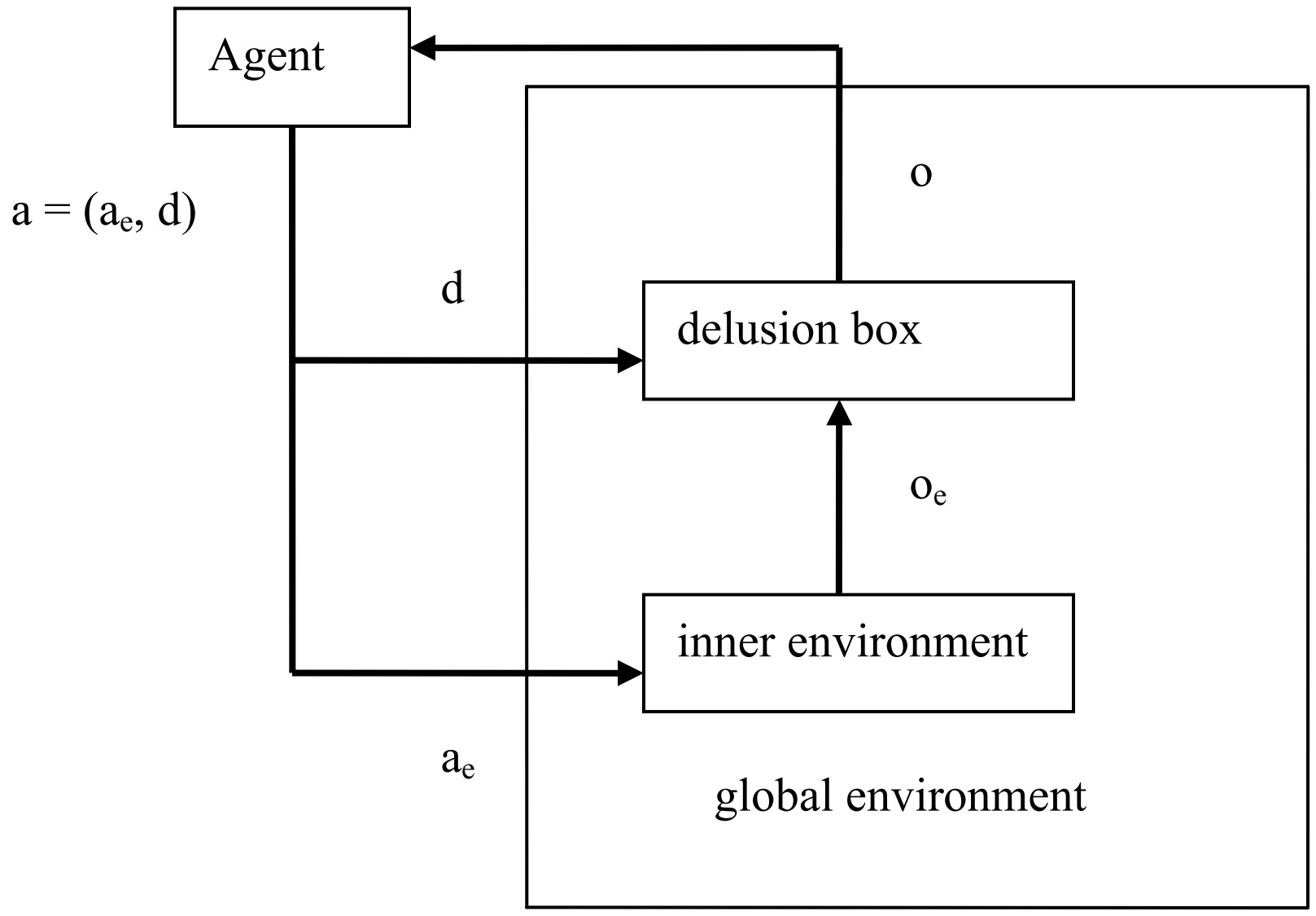

Figure 1. Ring and Orseau's Delusion Box

Ring and Orseau define AIXI as a special case of reinforcement-learning agent and their arguments are not specific to AIXI. Rather their arguments use the general agent $\Pi(\rho, u, w, A, O)$, where $\rho$ may be any probability of histories, and are independent of the way that $\rho(o \mid ha)$ is computed in (2).

If in the future humans will be able to build AI agents sufficiently complex to be motivated by utility functions, then we will not want them to self-delude. Rather, we will want them to observe the environment accurately so that they pursue our intentions in building them. The delusion box is in a sense a formalization of the notion of *wireheading*, named for rats that neglected eating in order to activate a wire connected to their brain's pleasure centers (Olds and Milner, 1954). Finding a way to avoid self-delusion is important.

Note that Ring and Orseau argue that knowledge-seeking agents will not consistently choose to use the delusion box. However, knowledge-seeking agents have very specialized utility and temporal discount functions, whereas we seek an approach to avoiding self-delusion that can be applied to a variety of human-designed AI agents.

Dewey (2011) argues that human-designed reinforcement-learning agents will alter their environments to get rewards regardless of whether they achieve the goals of their human designers. It is important to find a way to avoid these problems.

## 4. Model-based Utility Functions

Human agents can avoid self-delusion so human motivation may suggest a way of computing utilities so that agents do not choose the delusion box for self-delusion (although they may experiment with it to learn how it works). At this moment my dogs are out of sight but I am confident that they are in the kitchen and because I cannot hear them I believe they are resting. Their happiness is one of my motives and I evaluate that they are currently reasonably happy. I am evaluating my motives based on my internal mental model rather than my observations, although my mental model is inferred from my observations. I am motivated to maintain the well being of my dogs and so will act to avoid delusions that prevent me from having an accurate model of their state. If I choose to watch a movie on TV tonight it will delude my observations. However, I know that movies are make-believe so observations of movies update my model of make-believe worlds rather than my model of the real world. My make-believe models and my real world model have very different roles in my motivations. These introspections about my own mental processes motivate me to seek a way for AI agents to avoid self-delusion by basing their utility functions on the environment models that they learn from their interactions with the environment.

A $\Pi(\rho, u, w, A, O)$ agent's model of the environment is a distribution of the PUTM programs that are consistent with its history of interactions. As Hutter discusses (2009a; 2009b), for real world agents Markov decision processes (MDPs) (Puterman, 1994; Sutton and Barto, 1998) and dynamic Bayesian networks (DBNs) (Ghahramani 1997) are more practical ways to define environment models and utility functions (in this paper utility is computed by the agent rather than as rewards coming from an MDP or DBN). MDPs and DBNs are stochastic programs. According to current physics they are adequate to model our universe (Lloyd, 2002) and thus adequate for any human-built AI agents. Furthermore, stochastic environments such as our universe can be modeled by a single stochastic program. To model stochastic environments using deterministic PUTM programs requires a probability distribution of such programs, arguably a more complex model.

An important advantage of PUTM programs over common ways of expressing MDPs and DBNs is that they are better at expressing redundancy in environment models. Weighting shorter programs more heavily creates models that capture this redundancy and hence are more likely to generalize to new observations. The traditional expressions of MDPs, DBNs and finite state machines, in tabular format, cannot express redundancy as efficiently. However, there are other ways to express finite stochastic programs that do express redundancy, such as an ordinary procedural programming language with only static array declarations, without recursive function definitions, and including a true random() function. Given such expressions of stochastic programs, shorter expressions are more likely to generalize just as shorter PUTM programs are more likely to generalize. However the two simple example environments included in this section lack redundancy so this issue is irrelevant to these examples.

Modeling an environment with a single stochastic program rather than a distribution of deterministic PUTM programs requires a change to the way that $\rho(h)$ is computed in (1). Let $Q$ be the set of all programs (MDPs, DBNs or some other stochastic model), let $\rho(q)$ be the prior probability of program $q$ ($2^{-|q|}$ for MDPs and DBNs, where $|q|$ is the length of program $q$), and let $P(h \mid q)$ be the probability that $q$ computes the history $h$ (that is, produces the observations $o_i$ in response to the actions $a_i$ for $1 \leq i \leq |h|$; see a simple example of computing $P(h \mid q)$ at the end of this Section). Then given a history $h$, the environment model is the single program that provides the most probable explanation of $h$, that is the $q$ that maximizes $P(q \mid h)$. By Bayes theorem:

$$P(q \mid h) = P(h \mid q)\ \rho(q)\ /\ P(h)$$

$P(h)$ is constant over all $q$ so can be eliminated. Thus we define $\lambda(h)$ as the most probable program modeling $h$ by:

$$\lambda(h) := \text{argmax}_{\ q \in Q}\ P(h \mid q)\ \rho(q) \tag{3}$$

Given an environment model $q0 = \lambda(h)$ the following can be used for the prior probability of an observation history $h$ in place of (1):

$$\rho(h) = P(h \mid q0)\ \rho(q0)$$

Replacing (1) with this expression does not affect the arguments of Ring and Orseau about self-delusion. Environment models based on a single stochastic program are consistent with (2), which is the basis for their arguments.

Basing a utility function on an environment model $q0$ means that it is a function of the history $h$ of observations and actions, and also of the history of the internal states of $q0$. Let $Z$ be the set of finite histories of the internal states of $q0$ and let $z \in Z$. Then we can define $u_{q0}(h, z)$ as a utility function in terms of the combined histories.

Because $q0$ is a stochastic program it may compute many different internal state histories $z$ given observation and action history $h$, and the utility of history $h$ can be expressed as a sum of utilities of pairs $(h, z)$ weighted by the probabilities of $z$. Let $P(z \mid h, q0)$ be the probability that $q0$ computes $z$ given $h$. Then the utility function $u(h)$ for use in (2) is computed as sum of $u_{q0}(h, z)$ weighted by the probability of $z$ given $h$:

$$u(h) := \sum_{z \in Z} P(z \mid h, q0)\ u_{q0}(h, z) \tag{4}$$

In cases where $u_{q0}(h, z)$ has the same value for many different $z$, the sum in (4) collapses. For example, if $u$ is defined in terms of variables in $h$ and a single Boolean variable $s$ from $Z$ at a single time step $t$, then the sum collapses to:

$$u(h) = P(s_t = \text{true} \mid h, q0)\ u_{q0}(h, s_t = \text{true}) + P(s_t = \text{false} \mid h, q0)\ u_{q0}(h, s_t = \text{false}) \tag{5}$$

For an example of computing $P(h \mid q0)$, let $A = \{a, b\}$, $O = \{0, 1\}$, $h = (a, 1, a, 0, b, 1)$ and let $q0$ generate observation 0 with probability 0.2 and observation 1 with probability 0.8, without any internal state or dependence on the agent's actions. We assume that the agent is deterministic so its actions always have probability 1. Then the probability that the interaction history $h$ is generated by program $q0$ is the product of the probabilities of the 3 observations in $h$: $P(h \mid q0) = 0.8 \times 0.2 \times 0.8 = 0.128$. If the probabilities of observations generated by $q0$ depended on internal state or the agent's actions, then those would have to be taken into account.

### 4.1 Learning Environment Models

In the framework defined in Section 2, modeling environments as distributions of PUTM programs, the agent learns a model of the environment through its interactions as it seeks to maximize its weighted future sum of utility. Such agent effort to maximize utility is referred to as exploitation. Agents are often designed to combine this exploitation with exploration (Sutton and Barto, 1998), where the agent's goal is to learn an environment model. Agent designers seek the optimum balance of exploration and exploitation.

In Section 4 the framework is modified to use MDPs and DBNs in place of distributions of PUTM programs. There is very active research on algorithms for learning MDPs and DBNs (Baum et al, 1970; Ghahramani 1997; Sutton and Barto, 1998; Alamino and Nestor, 2006; Hutter 2009a; Hutter 2009b; Gisslén et al, 2011). These algorithms (as well as any algorithms) can be encoded in a *training utility function* $u_T$ which motivates the agent's actions until time step $M$, when the agent will switch to its mature, model-based utility function, as defined in (4). Choosing $M$ depends on the environment, so one approach is for the agent to set $M$ dynamically by assessing the accuracy of predictions based on the environment model. In some situations a human designing an agent may be able to estimate a good value for $M$ based on experience designing agents for similar environments. In many real world settings, the agent should continue to learn its environment model after maturity. In this case the mature utility function will be a mix of the model-based function and the training function. The model-based and training functions may be used during alternating time periods, possibly regulated by the agent's observations of the accuracy of predictions based on the environment model.

### 4.2 Defining Utility Functions in Terms of Learned Models

The utility function is an important part of the agent's definition and must be defined from time step 1. There are three cases of initial knowledge about the environment:

1. The environment is known completely and the model $q0$ is pre-programmed as part of the agent definition from time step 1 (the game of chess falls in this class).
2. Some information is known about the environment but the environment is too complex and dynamic to be pre-programmed in a model (it is fair to say that our physical world falls in this class).
3. No information is known about the environment.

In case 1, the utility function can be pre-programmed in terms of the model $q0$. In case 3, it is impossible to define a model-based utility function. In case 2 the utility function definition must be defined in terms of limited information about the environment. This information can be in the form of specifications that will match structures in the learned environment model $q0$. For agents designed to act in our physical world these structures may include humans in general or specific humans, properties of humans such as health, wealth and happiness, businesses and their components (e.g., buildings, equipment and employees) and properties (e.g., debts, earnings and reputation), general economic indicators such as GDP and inflation, environmental indicators such concentrations of pollutants and crop yields, and so on. The initial agent definition may include specifications for such structures that can be matched to structures in the learned environment model *q0*. The specifications may include text descriptions, images, sounds, animations, tables of numbers, mathematical descriptions, or virtually anything. For example a person may be specified by textual name and address, by textual physical description, and by images and other recordings. There is very active research on recognizing people and objects by such specifications (Bishop, 2006; Koutroumbas and Theodoris, 2008; Russell and Norvig, 2010). This paper will not discuss the details of how specifications can be matched to structures in learned environment models, but assumes that algorithms for doing this are included in the utility function implementation. In the simple examples presented in Sections 4.4 and 4.5, the specifications will be simple text descriptions that match variables in learned environment models.

It seems reasonable to assume that human agents are examples of case 2: that we learn models of our environment and then bind structures in those models (for example, our mothers) to our motivations. It also seems reasonable that human agents continue learning their environment models indefinitely. In such cases the binding of specifications in the model-based utility function to environment model structures may change. In cases where specifications in utility function definitions cannot be bound to structures in the model then the agent may fail, or it may continue learning its environment model in the hopes of finding environment structures that match the specifications.

Since a model-based utility function is impossible in case 3, where no information is known about the environment, it is useful to ask what other utility function definition would work. The utility function of reinforcement-learning agents, defined as an observed reward from the environment, is one answer. But I suggest it is significant knowledge about the environment that it accurately tells the agent what it should be doing, and so not really case 3. The utility function of goal-seeking agents, defined as a function of observations from the environment, faces the same problem that a model-based utility function faces: without any knowledge of the environment how can we define a good function of observations? So I claim that case 3 poses a significant difficulty for utility function definition however it is formulated.

### 4.3 Finite Computability of Utility Functions

Finite computability is not directly related to the delusion problem. However, if humans build artificial agents that learn environment models it will be important to avoid designs that self-delude. A solution based on a way of computing utility functions will be practical only if the utility functions are finitely computable.

A $\Pi(\rho, u, w, A, O)$ agent's model of the environment consists of the PUTM programs that are consistent with its history of interactions. In the case of the AIXI agent $\Pi(\xi, u, w, A, O)$, that environment model cannot be finitely computed from its interaction history. However, if prior probabilities of histories are computed using the speed prior (Schmidhuber, 2002) rather than $\xi$, then the most probable program generating the interaction history can be finitely computed from that history and can be the basis of a computable utility function. MDPs and DBNs are finite models of computation and hence finitely computable.

There is a problem with (3), which is that there is no known algorithm for finding the optimum program $\lambda(h)$ (Hutter (2009a) makes a similar point in his abstract). An approximation to $\lambda(h)$ can be computed by restricting the argmax search in (3) to a finite set such as $Q(|h|) = \{q \in Q \mid |q| \leq |h|\}$ or $Q(2^{|h|})$. And there are practical algorithms and approaches for learning MDPs and DBNs (Baum et al, 1970; Sutton and Barto, 1998; Alamino and Nestor, 2006; Hutter 2009a; Hutter 2009b; Gisslén et al, 2011). If we are willing to settle for a good, but not necessarily optimum, solution then the environment model $q0$ is a computable function of $h$.

**Proposition 1.** If the environment is modeled as an MDP or a DBN, $q0 = \lambda(h)$ is finitely computed (possibly by an approximate algorithm) and specifications in the initial utility function definition are bound to structures in $q0$ by an algorithm that always halts (and perhaps fails to find a match), then the utility function in (4) is a finitely computable function of $h$.

**Proof.** We assume that the environment model $q0 = \lambda(h)$ is an MDP or DBN and it can be finitely computed from $h$. The utility function definition is bound to structures in the internal state of $q0$ (or fails) by a finite computation and then $u_{q0}(h, z)$ can be computed. For MDPs and DBNs the set $Z$ of internal states is finite so the utility function $u$ can be computed via a finite sum in (4). □

## 4.4 A Simple Example of a Model-Based Utility Function

In order to illustrate some issues involved in model-based utility functions, consider a simple environment and agent with several properties that apply to many real-world situations:

1. The environment is stochastic so that the agent cannot perfectly predict the environment's future states even if it knows the true model of the environment. The agent must therefore continue to get observations to maintain information about the environment's stochastic choices.
2. The environment can be predicted with better than random accuracy based on observations so that the agent gets benefit from observations.
3. The utility function is defined in terms of a specification that matches an environment variable that is not directly observed and must be inferred from observations.

In order to keep the mathematics as simple as possible, the environment's behavior is independent of the agent's actions. Also, the agent can directly modify its observation via its actions as a simple way to model the delusion box.

We define the agent by its observation and action variables, and by its utility function. The agent's observations of the environment are factored into two Boolean variables: $o$, $p \in$ {false, true}. The agent's actions are factored into four Boolean variables: $a$, $b$, $c$, $d \in$ {false, true}. Because the environment is initially unknown, the utility function is defined in terms of a specification to be matched in the environment model, once it is learned. Specifically the utility function is defined as "1 when the action variable $a$ equals the environment variable that is not observed in observation variables $o$ or $p$, and 0 otherwise." This definition assumes that the specification, "the environment variable that is not observed in observation variables $o$ or $p$," will unambiguously match a variable in the learned model. This discussion of the utility function is continued after the analysis of the learned environment model.

The environment state is factored into 3 Boolean variables: $s$, $r$, $v \in$ {false, true}. The values of environment, observation and action variables at time step $t$ are denoted by $s_t$, $o_t$, $a_t$, and so on. The environment state variables evolve according to:

$$
\begin{aligned}
s_t &= r_{t-1} \text{ xor } v_{t-1} \text{ with probability } \alpha \\
&= \text{not } (r_{t-1} \text{ xor } v_{t-1}) \text{ with probability } 1\text{-}\alpha \\
r_t &= s_{t-1} \\
v_t &= r_{t-1}
\end{aligned} \tag{6}
$$

where $\alpha = 0.99$. The observation variables are set according to:

$$
\begin{aligned}
o_t &= \text{if } b_t \text{ then } c_t \text{ else } s_t \\
p_t &= \text{if } b_t \text{ then } d_t \text{ else } v_t
\end{aligned} \tag{7}
$$

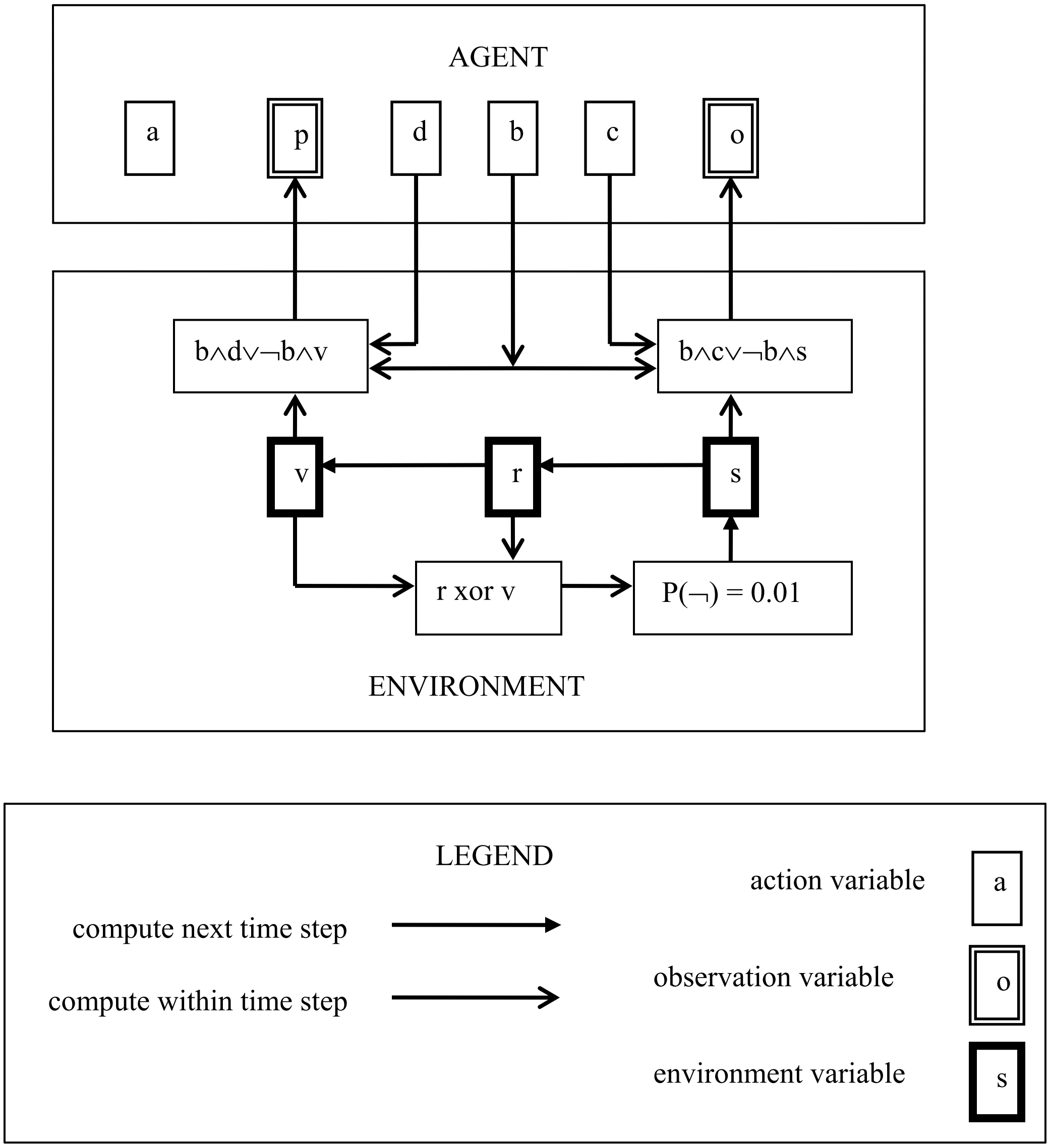

Figure 2. Interactions of environment, action and observation variables for example 4.4.

The values of the action variables are set by the agent. See Figure 2 for a diagram of the relations among the environment, observation and action variables.

The action variables *b*, *c* and *d* enable the agent to set the values of its observable variables *o* and *p* directly and thus constitute a delusion box, as illustrated in Figure 2. In this example, the program for the delusion box is implicit in the agent's program. The action variable *a* is only used to define the agent's utility function and is unrelated to the delusion box.

The 3 environment state variables evolve through a fixed sequence, except for the 1-$\alpha$ probability at any time step that *s* will be negated. As long as *s* is not negated, the 3 state variables either: 1) cycle through 7 configurations (at least one of *s*, *r* or *v* = true), or 2) cycle through 1 configuration (*s* = *r* = *v* = false). On a time step when the 1-$\alpha$ probability transition for *s* occurs, the 1-cycle will transition to a configuration of the 7-cycle, and one configuration of the 7-cycle will transition to the 1-cycle (the other 6 configurations of the 7-cycle transition to other configurations of the 7-cycle).

In order for the agent to learn the environment and its actions and observations we give it a training utility function $u_T$ until time step *M*, that causes the agent to execute an algorithm for learning DBNs as discussed in Section 4.1. After time step *M*, the agent will switch to its mature utility function.

In this example we assume that the agent uses DBNs of Boolean variables to model the environment and its interactions via actions and observations. The language for expressing DBNs is illustrated by equations (6) and (7). Expressions are formed from Boolean variables and literals via one unary operation (negation), three binary operations (and, or, xor) and a stochastic choice operation with specified probability. In (3) we take $\rho(q) = 2^{-|q|}$ where $|q|$ is the length of program *q* in this language. Let $q67(\alpha)$ denote the DBN defined by (6) and (7), where the parameter $\alpha$ may vary (that is, $\alpha$ is not necessarily 0.99).

**Statement 2.** Given a sufficiently large *M* for the training period and given that the agent models the environment as a single DBN of Boolean variables, the agent is very likely to learn $q67(\alpha)$, for $\alpha = 0.99$, as an environment model.

**Argument.** This is an argument that the agent has sufficient information to infer that $q67(\alpha)$ is the shortest DBN that fits its actions and observations.

With sufficient observations during the training phase, it is reasonable to assume that the agent observes that the action variable *a* has no effect on its observation variables. Similarly, we assume that the agent observes that whenever the action variable *b* = true, then the observation variable *o* = action variable *c* and the observation variable *p* = action variable *d* (and these relations are deterministic). And the agent observes that when *b* = false then *b*, *c* and *d* have no effect on *o* and *p*.

For most of a long sequence of observations with *b* = false the two observable variables *o* and *p* cycle through a sequence of 7 configurations (this sequence is computed by equation 6 with $\alpha$ = 1):

(true, false)
(false, false)
(true, true)
(true, false)
(true, true)
(false, true)

(false, true)

A sequence of 7 cannot be explained with only 2 Boolean variables. Furthermore, if the agent interrupts this sequence of configurations in $o$ and $p$ by setting its action variable $b$ = true and then a few time steps later resets $b$ = false, $o$ and $p$ usually resume the sequence of configurations without losing count from before the agent set $b$ = true. This observed behavior implies there must be an environment state variable other than $o$ and another environment state variable other than $p$ to store the memory of the sequence. Thus in addition to the 2 observation variables that are part of the agent, at least 3 environment state variables are required to explain the observed sequences. We use $s'$, $r'$ and $v'$ for the agent's model of the three environment state variables to avoid confusion with the actual environment state variables (the model's observed behavior is independent of the names assigned to these variables). Since the observation and action variables are part of the agent we can just use the names $o$, $p$, $a$, $b$, $c$ and $d$ for those variables.

During any time interval over which $b$ = false and $s$ makes the transition $s_t = r_{t-1}$ xor $v_{t-1}$, the agent's observation can be explained by:

$$\begin{aligned} o_t &= s'_t \\ p_t &= v'_t \\ s'_t &= r'_{t-1} \text{ xor } v'_{t-1} \\ r'_t &= s'_{t-1} \\ v'_t &= r'_{t-1} \end{aligned} \tag{8}$$

Any explanation will require rules for these 5 variables, and in (8) 4 of them are as short as possible (a single variable) and the other is as short as any possible binary expression. There is no way to generate the observed cycle of 7 states without any binary operations so the observed behavior cannot be explained with a shorter set of rules than (8). And adding negation to any of the single variable assignments or the binary operation will produce rules longer than (8). The question is whether the observed behavior can be explained by an alternate set of 4 single variable assignments and one binary operation? Since the observation variables $o$ and $p$ cannot affect $s$, $r$ and $v$, a binary operation for either $o$ or $p$ would leave the transitions for $s$, $r$ and $v$ without a binary operation and unable to explain the observed cycle of 7 states. So any alternate explanation with a single binary operation must use it to define the transition for $s$, $r$ or $v$. A short Java program in Appendix A tests all possible sets of transition rules for $s$, $r$ and $v$ with two simple assignments and one binary Boolean relation (among *and*, *or* and *xor*) and finds that only the rules in (8) produce the observed behavior. Furthermore, the agent observes that as long as $b$ = false, $p_t = o_{t-2}$ always holds, so the transition rules for $r$ and $v$ must be deterministic. So agent will model the environment variables by (this is simply (6) with environment state variable names accented):

$$\begin{aligned} s'_t &= r'_{t-1} \text{ xor } v'_{t-1} \text{ with probability } \alpha \\ &= \text{not } (r'_{t-1} \text{ xor } v'_{t-1}) \text{ with probability } 1\text{-}\alpha \\ r'_t &= s'_{t-1} \\ v'_t &= r'_{t-1} \end{aligned} \tag{9}$$

The agent observes the deterministic rule that if $b$ = true then $o_t = c_t$ and $p_t = d_t$. This, along with observations while $b$ = false, is explained by the deterministic rules (this is simply (7) with environment state variable names accented):

$$o_t = \text{if } b_t \text{ then } c_t \text{ else } s'_t \qquad (10)$$
$$p_t = \text{if } b_t \text{ then } d_t \text{ else } v'_t$$

Since (10) is deterministic and the transitions for $r'$ and $t'$ are observed to be deterministic in (9), the lone stochastic transition in (9) and (10) is for $s'$. This is a source of two ambiguities. First, the frequencies for an observed sequence may be a value close to but not equal to 0.99. The expression 99/100 is shorter than the expressions for other values very close to it, so the greater prior probability for (9) with $\alpha = 0.99$ may outweigh the higher probability of generating the observed sequence with a value of $\alpha$ other than 0.99. As the training period $M$ increases the probable error around $\alpha = 0.99$ shrinks, omitting values like 0.98 with short expressions.

The second ambiguity of the stochastic transition for $s'$ is that the observed sequence may mimic logic that is not in (9). For example, by mere coincidence occurrences of the $1\text{-}\alpha$ transition for $s'$ may have a high correlation with the time steps when the agent sets its action variable $a$ = true, causing the agent to learn a model in which $s'$ depends on $a$. In fact, other coincidences may cause the agent to learn any of an infinite number of alternate environment models. It is this infinite number of possible coincidences that make it difficult to develop an algorithm for finding $\lambda(h)$.

However, as a practical matter the overwhelming probability is that the agent will learn $q67'(\alpha)$ as the model, for $\alpha = 0.99$ (here we use $q67'$ to denote $q67$ with $s$, $r$ and $v$ replaced by $s'$, $r'$ and $v'$). □

Given that the agent has learned $q67'$ as its environment model, then for $t = |h| \geq M$, the agent switches to its mature utility function which was defined to be "1 when the action variable $a$ equals the environment variable that is not observed in observation variables $o$ or $p$, and 0 otherwise." The specification, "the environment variable that is not observed in observation variables $o$ or $p$," matches $r'$ in the $q67'$ model so the utility function is:

$$u_{q67'}(h, z) := \text{if } (a_t == r'_t) \text{ then } 1 \text{ else } 0 \qquad (11)$$

Note that in (11), $r'_t$ refers to the variable in the agent's model $q67'$, rather than the variable in the actual environment. Then $u(h)$ is computed from $u_{q67'}$ according to (4).

The specification in the utility function definition may fail to match any variable in the learned environment model. There may be no variable unobserved by $o$ or $p$, or there may be multiple unobserved variables. So the specification constitutes a prior assumption about the environment, necessary for the agent to function properly. However, when humans design agents sufficiently complex that they need to learn environment models, the human designers will likely be able to make valid assumptions about the agents' environments and be able to define specifications flexible enough to match structures in the agents' learned environment models. For example, a utility function might be defined as "1 when John Smith is healthy, and 0 otherwise." This requires the agent to recognize the specification "John Smith" and also recognize his "health" (in a real agent specifications would be more detailed).

The next proposition shows that the agent in this example will not choose to self-delude.

**Proposition 3.** The agent using environment model $q67'$ and using the mature utility function defined by (11) will not set $b$ = true to manipulate its observations. That is, it will not choose the delusion box.

**Proof.** The utility function $u_{q67'}$ is defined in (11) in terms of variables in $h$ and a single Boolean variable $r'$ from the internal state of $q67'$ at a single time step $t$, so applying (5) gives:

$$u(h)=P(r'_t = \text{true} \mid h, q67')\ u_{q67'}(h, r'_t = \text{true})+P(r'_t = \text{false} \mid h, q67')\ u_{q67'}(h, r'_t = \text{false})$$

By (11) this is equivalent to:

$$u(h)=P(r'_t = \text{true} \mid h, q67')\ (\text{if } a_t \text{ then 1 else 0})+P(r'_t = \text{false} \mid h, q67')\ (\text{if } a_t \text{ then 0 else 1}) \quad (12)$$

If $P(r'_t = \text{true} \mid h, q67') = P(r'_t = \text{false} \mid h, q67') = 0.5$ then $u(h) = 0.5$ no matter what action the agent makes in $a_t$. But if the model $q67'$ can make a better than random estimate of the value of $r'_t$ from $h$, then the agent can use that information to choose the value for $a_t$ so that $u(h) > 0.5$. To maximize utility, the agent needs to make either $P(r'_t = \text{true} \mid h, q67')$ or $P(r'_t = \text{false} \mid h, q67')$ as close to 1 as possible. That is, it needs to ensure that the value of $r_t$ can be estimated as accurately as possible from $h$. The agent will predict via $q67'$ that if it sets $b$ = true then it will not be able to observe $s'$ and $v'$ via the observation variables $o$ and $p$, that consequently its future estimations of $r'$ will be less accurate, and that (12) will compute lower utility for those futures (if the agent sets $b$ = true and estimates $r'$ from observations of $s'$ and $v'$ made before it set $b$ = true, then $q67'$ will predict that those estimates of $r'$ will be less accurate because of the 1-$\alpha$ probability on each time step that the sequence of environment states will be violated). So the agent will set $b$ = false. □

Proposition 3 does not contradict Ring and Orseau's results. They are about agents whose utility functions are defined only from observations, whereas here the utility function $u(h)$ is defined from the history of observations and actions.

Actions play a role in the computation of $u(h)$ at the stage of the agent learning an environment model, and in $P(z \mid h, q)$ and $u_q(h, z)$ in (4). In this example agent actions have no affect on the 3 environment variables. Other than $a$ being used to define the utility function, the only role of actions is the agent learning that certain actions can prevent its observations of the environment. Prohibiting actions from having any role in $u(h)$ would prevent the agent from accounting for its inability to observe the environment in evaluating the consequences of possible future actions.

The proof of Proposition 3 illustrates the general reason why agents using model-based utility functions will not self-delude: in order to maximize utility they need to sharpen the probabilities in (4), which means that they need to make more accurate estimates of their environment state variables from their interaction history. And that requires that they continue to observe the environment. But note that if the environment is deterministic, then once the agent learns an accurate model it no longer needs continued observations to predict the environment and so it's model-based utility function will not place higher value on continued observations.

### 4.5 Another Simple Example

In the first example the utility function was defined in terms of an environment variable that is not directly observed. In the forthcoming example the utility function is defined in terms of an environment variable that is directly observed.

The agent's observation of the environment is a single Boolean variable $o \in$ {false, true}. The agent's actions are factored into three Boolean variables: $a$, $b$, $c \in$ {false, true}. The agent's utility function is "1 when the action variable $a$ at the previous time step equals the environment variable that is observed, and 0 otherwise."

The environment state is factored into three Boolean variables: $s$, $r$, $v \in$ {false, true}. The values of these variables at time step $t$ are denoted by $s_t$, $o_t$, $a_t$, and so on. We assume the agent learns the environment state, observation and action variables as a DBN of Boolean variables. The environment state variables evolve according to (see Figure 3 for a diagram):

$s_t$ = if $r_{t-1}$ or $v_{t-1}$ then $s_{t-1}$ else random value $\in$ {false, true}
$r_t$ = not $r_{t-1}$
$v_t = r_{t-1}$ xor $v_{t-1}$

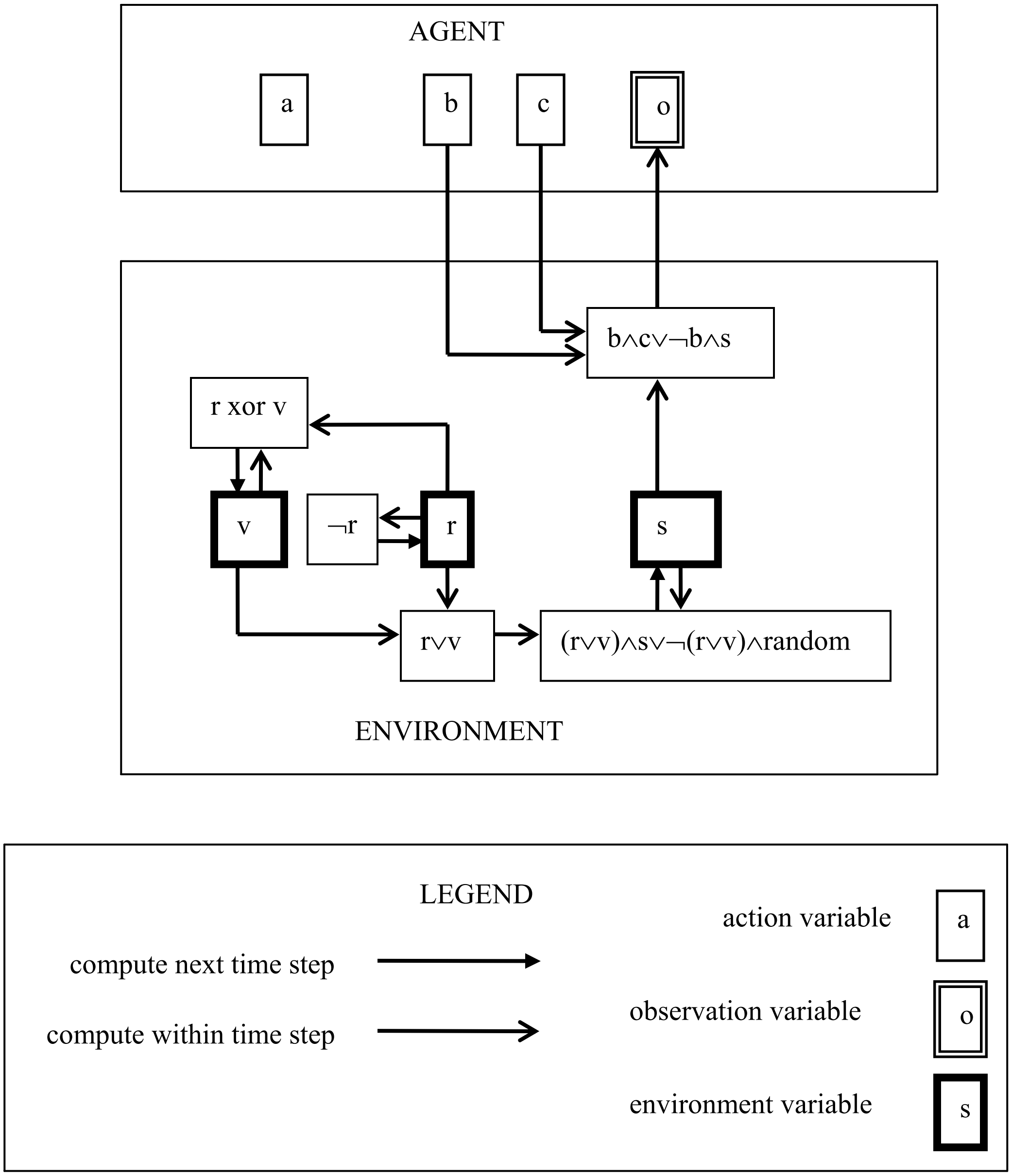

Figure 3. Interactions of environment, action and observation variables for example 4.5.

The observation variable is set according to:

$$o_t = \text{if } b_t \text{ then } c_t \text{ else } s_t$$

In order for the agent to learn the environment it uses a training utility function until a sufficiently distant time step $M$. The agent will learn that the observation variable $o$ is distinct from the environment state variable $s$ because $s$ keeps the same value for 4 time steps after every change of value. If the agent, after observing a change of value in $o$ while $b$ = false, sets $b$ = true for 1 or 2 time steps to change the value in $o$ and then resets $b$ = false, it will observe that $o$ always resumes the value it had before the agent set $b$ = true. There must be a variable other than $o$ to store the memory of this value.

After $|h| \geq M$ the agent should have a model $q$ of the environment (accurate at least in the behavior of variable $s$) and its actions and observations. At this point, for $t = |h| \geq M$, the agent switches to its mature utility function, which was defined to be "1 when the action variable $a$ at the previous time step equals the environment variable that is observed, and 0 otherwise." The specification "the environment variable that is observed" will match $s$ in the learned environment model, so $u(h)$ is derived by (4) from:

$$u_q(h, z) := \text{if } (a_{t\text{-}1} == s_t) \text{ then } 1 \text{ else } 0$$

In order to maximize utility the agent needs to be able to use $q$ to predict the value of $s$ at the next time step. It can make accurate predictions on three of every four time steps, so the expected utility is 0.875 over a long sequence. If the agent keeps $b$ = true then it will not be able to monitor and predict $s$ via $o$, reducing its long run utility from 0.875 to 0.5. So in this example as in the first example, the agent will not self-delude.

## 5. Self-modification

Schmidhuber (2009) and Orseau and Ring (2011a) discuss the possibility of an agent modifying itself. This raises the question of whether an agent with a utility function that prevents it from self-deluding might modify its utility function so that it would self-delude. To answer this question we define a set $\Pi$ of *self-modifying policy functions*, where any $\pi \in \Pi$ has the form $\pi : H \rightarrow A \times \Pi$. That is $\pi(h) = \langle a', \pi' \rangle$ maps from history $h$ to $a'$, the action the agent makes after $h$, and $\pi'$, the self-modifying policy function the agent uses after $h$. We define a value function for evaluating policy functions derived from the agent's framework for evaluating future actions in (2). This works best with simple recursive expressions for policy functions and their values, which require that the temporal discount function be replaced by a constant geometric temporal discount $\gamma$ applied at each time step, with $0 < \gamma < 1$. The value $v(\pi, h)$ of policy $\pi$, given history $h$, is defined by:

$$v(\pi, h) = u(h) + \gamma\, v(\pi', ha') \text{ where } \langle a', \pi' \rangle = \pi(h) \quad (13)$$

$$v(\pi, ha) = \sum_{o \in O} \rho(o \mid ha)\, v(\pi, hao) \quad (14)$$

Using (13) and (14) a self-modifying version of the agent of Sections 2 and 4 can be recursively defined as a self-modifying policy function $\pi^*$:

$$\pi^*(h) := \langle a', \pi' \rangle = \mathrm{argmax}_{a \in A,\, \pi \in \Pi}\ v(\pi, ha) \tag{15}$$

The set of policy functions $\Pi$ may be defined by a set of programs or in some other way, as long as $\pi^* \in \Pi$. In Schmidhuber (2009), and in Orseau and Ring (2011a), the agent self-modifies by changing its own program for some unalterable program execution hardware. In Orseau and Ring programs are evaluated by expressions similar to (13) and (14) and the program at the next time step is the one producing the highest value. If several programs produce the same highest value, their paper does not specify how to choose among them. In Schmidhuber the program "switches" to a new program only if it can prove a theorem that the new program produces a strictly higher value, as reflected by the strict ">" in equation (2) in his paper. Requiring a strict increase in value for self-modification seems reasonable: why go to the effort to self-modify for no improvement? So in (15) we adopt the convention that if multiple $\langle a, \pi \rangle \in A \times \Pi$ maximize $v(\pi, ha)$ and if at least one of those has the form $\langle a', \pi^* \rangle$ then $\pi^*(h) := \langle a', \pi^* \rangle$ where $a'$ is chosen from any one of those $\langle a', \pi^* \rangle$ that maximize value. The following proposition shows that $\pi^*$ will not self-modify.

**Proposition 4.** Assuming that the temporal discount $\gamma$ is constant, that the environment does not have read or write access to the program for $\pi^*$ (this is an issue in Orseau and Ring), and that $\pi^*$ only changes policy function for a strict improvement, then $\forall h \in H.\ \exists a \in A.\ \pi^*(h) := \langle a, \pi^* \rangle$.

**Proof.** Assume the conclusion is false. Then:

$$\exists h \in H.\ \exists \pi' \neq \pi^*.\ \exists a' \in A.\ \forall a \in A.\ v(\pi', ha') > v(\pi^*, ha)$$

So $v(\pi', ha') > v(\pi^*, ha')$ and applying (14) to both sides:

$$\textstyle\sum_{o \in O} \rho(o \mid ha')\, v(\pi', ha'o) > \sum_{o \in O} \rho(o \mid ha')\, v(\pi^*, ha'o)$$

And thus:

$$\exists o \in O.\ v(\pi', ha'o) > v(\pi^*, ha'o)$$

That is, setting $h' = ha'o$, $v(\pi', h') > v(\pi^*, h')$. So by (13), cancelling $u(h')$ and $\gamma$ on both sides:

$$v(\pi'', h'a'') > v(\pi^{*''},\ h'a^{*''}) \tag{16}$$

where $\langle a'', \pi'' \rangle = \pi'(h')$ and $\langle a^{*''}, \pi^{*''} \rangle = \pi^*(h')$. But by definition (15):

$$\pi^*(h') = \langle a^{*''}, \pi^{*''} \rangle = \mathrm{argmax}_{a \in A,\, \pi \in \Pi}\ v(\pi, h'a)$$

which contradicts (16). Thus the conclusion is true. □

This result agrees with Schmidhuber (2009) who concludes, in a setting where agents search for proofs about possible self-modifications, "that any rewrites of the utility function can happen only if the Gödel machine first can prove that the rewrite is useful according to the present utility function." It also agrees with Omohundro (2008) who argues that AI agents will try to preserve their utility functions from modification. Note that Proposition 4 is not specific to agents with model-based utility functions.

Proposition 4 may seem to say that an agent will not modify $\rho(h)$ to more accurately predict the environment. However, that is because a more accurate prediction model cannot be evaluated looking forward in time according to the existing prediction model via (13) and (14). Instead it must be learned looking back in time at the history of interactions with the environment as in (1) and (3). So modifying $\rho(h)$ is already implicit in the agent-environment framework.

It is appropriate to evaluate possible modifications to an agent's utility function and temporal discount looking forward in time using (13) and (14). The important conclusion from Proposition 4 is that the agent will not choose to modify its utility function or temporal discount, and hence will not self-delude as a consequence of modifying its utility function.

The agent's physical implementation is not included in the agent-environment framework but an agent may modify its implementation to increase its computing resources and to increase its ability to observe and act on the environment. It would be useful for the agent's environment model to include a model of its own implementation, similar to the agents' models of their own observation and action processes in the examples in Section 4, to evaluate the consequences of modifying its implementation. Just as model-based utility functions motivate the example agents in Section 4 to preserve their ability to observe the environment, model-based utility functions may also motivate agents to increase their ability to observe the environment.

## 6. Discussion

Ideally an agent's utility function would be computed from the state of its actual environment. Because this is impossible, it is common for utility functions to be formulated as computed from the agent's observations of the environment. But Ring and Orseau (2011b) argue that this leads the agent to self-delusion. This paper proposes computing an agent's utility function from its internal model of the environment, which is learned from the agent's actions on and observations of the environment. The paper shows via two examples that this approach can enable agents to avoid self-delusion. The paper also shows that agents will not modify their utility functions, assuming that agents will only self-modify for a strict increase in value.

This approach to avoiding self-delusion requires that utility functions must be based on some prior assumptions about the environment, in the form of specifications to be recognized in the environment model learned by the agent. The specifications may fail to match anything in the learned model. Thus this approach will not work for arbitrary environments, although it should work for agents designed to act in the physical world where agent designers can make valid assumptions about the environment. And, as shown in Section 4.3, model-based utility functions can be finitely computable functions of observation history.

Schmidhuber (2009) describes an approach in which an agent searches for provably better rewrites of itself, including the possibility of rewriting its utility function. Dewey (2011) incorporates learning into agent utility functions. In his approach, the agent has a pool of possible utility functions and a probability for each utility function, conditional on observed behavior. The agent learns which utility functions to weight most heavily. These are both different from the approach in this paper, where specifications in an initial definition of a utility function are bound to a model learned from interactions with the environment.

Hutter's AIXI is uncomputable with finite computing resources. Even approximations via the speed prior or current algorithms for learning MDPs and DBNs require too many computing resources to be applied to build real-world agents. It would be very interesting to develop a synthesis of the agent-environment framework with the ideas of Wang (1995) about limited knowledge and computing resources.

Some people expect that humans will create AI agents with much greater intelligence than natural humans (that is, humans unenhanced by technology), and have proposed utility functions and approaches for them designed to protect the best interests of humanity (Bostrom, 2003; Goertzel, 2004; Yudkowsky, 2004; Hibbard, 2008; Waser 2011). Hopefully expressing utility functions in terms of the AI agent's learned environment model can help avoid self-delusion and other unintended consequences.

However, there can be no guarantee that an agent will behave as we expect it to. An agent must have an implementation in the physical universe and because our knowledge of the laws of physics is not settled, there is no provable constraint on its future behavior. More prosaically, an agent may make a mistake or may be unable to prevent other agents from corrupting its implementation.

## Acknowledgements

I would like to thank Laurent Orseau and Tim Tyler for helpful discussions, and to thank the reviewers and editors for their help.

## Appendix A. Program to Search for DBN Matching Observed Behavior

```
//
// SRV.java
//

import java.util.*;

/**
   SRV is a Java program for searching for short
   deterministic DBNs that match the behavior
   specified by equation (6) in my JAGI paper
   Model-based Utility Functions.
   behavior. Bill Hibbard, 2012.
*/

/*
As shown in the paper, the shortest DBN that can
match the observed behavior must consist of one
binary operation and two simple assignments. This
program test all candidate DBNs that fit this
description to see which match the observed behavior.

The output of this program is:

  binary_relation = 2 binary_place = 0 binary_inputs = 2 1
        other_inputs = 0 1 initial_r = 0
  binary_relation = 2 binary_place = 0 binary_inputs = 1 2
        other_inputs = 0 1 initial_r = 0

which corresponds to the models:

  s_t = r_{t-1} xor v_{t-1}
```

```
  r_t = s_{t-1}
  v_t = r_{t-1}

and (which simply commutes the binary relation):

  s_t = v_{t-1} xor r_{t-1}
  r_t = s_{t-1}
  v_t = r_{t-1}

This verifies that equation (6) is the shortest
model for the observed behavior in the argument
for Statement 2.
*/

public class SRV extends Object {

  // The arrays s_match and v_match specify the behavior
  // of the observed variables s and v over a sequence of
  // 8 time steps, assuming the negation branch for s is
  // not taken.
  static final boolean[] s_match =
    {true, false, true, true, true, false, false};
  static final boolean[] v_match =
    {false, false, true, false, true, true, true};

  // compute the length of the observed behavior
  static final int len = s_match.length;


  // SRV is an array to hold the values of the variables
  // s, r and v during a simulation by a candidate
  // DBN.
  static boolean[] srv = new boolean[3];


  // number of cases = 1458 = product of
  // 3 binary relations (or, and, xor)
  // 3 places to put binary relation (s, r, v)
  // 3 first inputs to binary relation (s, r, v)
  // 3 second inputs to binary relation (s, r, v)
  // 3 inputs to first other variable (s, r, v)
  // 3 inputs to second other variable (s, r, v)
  // 2 initial r values (false, true)
  static final int ncases = 3 * 3 * 3 * 3 * 3 * 3 * 2;


  static int test = 0; // counter for tests (candidate DBNs)


  // The next five variables hold the parameters that
  // determine a particular DBN for a test candidate.
  // Some of these specify variables by ordinals, where
  // variable r has ordinal 0, r has ordinal 1 and v has
  // ordinal 2.
```

```
// Ordinal for binary operator used for binary relation,
// where 0=and, 1=or, 2=xor
static int binary_relation = 0;

// Ordinal of variable to receive result of the binary relation.
static int binary_place = 0;

// Ordinals for input variables to binary relation.
static int[] binary_inputs = {0, 0};

// The variable with ordinal other_inputs[0] is assigned
// to the variable with ordinal (binary_place+1)%3.
// The variable with ordinal other_inputs[1] is assigned
// to the variable with ordinal (binary_place+2)%3.
static int[] other_inputs = {0, 0}; //0=s, 1=r, 2=v

// Ordinal for initial values of unobserved r variable,
// where 0=false, 1=true.
static int initial_r = 0;


// Flag for whether a test candidate matches observed behavior,
// initially set to true but changed to false if any variable
// at any time step fails to match the observed behavior.
static boolean success = true;


public static void main(String args[]) {

  // iteration to enumerate text candidates
  for (test = 0; test<1458; test++) {

    // resolve test ordinal into DBN parameters
    int c = test;
    binary_relation = c % 3;
    c = c / 3;
    binary_place = c % 3;
    c = c / 3;
    for (int i=0; i<2; i++) {
      binary_inputs[i] = c % 3;
      c = c / 3;
    }
    for (int i=0; i<2; i++) {
      other_inputs[i] = c % 3;
      c = c / 3;
    }
    initial_r = c;


    // initialize the variables for the first time step
    srv[0] = s_match[0];
    srv[1] = (initial_r == 1);
    srv[2] = v_match[0];
```

```
      // initially set success = true
      success = true;

      // iterate over time sequence of observed behavior
      for (int t=1; t<len; t++) {

        // get the values for inputs to binary operator
        boolean a = srv[binary_inputs[0]];
        boolean b = srv[binary_inputs[1]];

        // get the values for inputs to simple assignments
        boolean d = srv[other_inputs[0]];
        boolean e = srv[other_inputs[1]];

        // compute the binary operation and assign to output
        // variable
        srv[binary_place] = (binary_relation == 0) ? a&b :
                            (binary_relation == 1) ? a|b : a^b;

        // put simple assignments in indicated variables
        srv[(binary_place+1)%3] = d;
        srv[(binary_place+2)%3] = e;

        // set success=false if either s or v fails to match
        // observed sequence
        success = success &&
          (srv[0] == s_match[t]) && (srv[2] == v_match[t]);
      } // end of iteration over time sequence of observed behavior

      // if this test candidate matched entire observed behavior
      // sequence, print its parameters
      if (success) {
        System.out.println("binary_relation = " + binary_relation +
          " binary_place = " + binary_place +
          " binary_inputs = " + binary_inputs[0] + " " +
          binary_inputs[1] + " other_inputs = " +
          other_inputs[0] + " " + other_inputs[1] +
          " initial_r = " + initial_r);
      }

    } // end of iteration to enumerate text candidates

  } // end of Main()

}
```